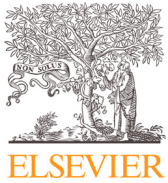
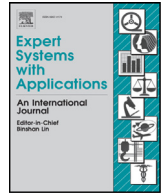

# Incremental personalized E-mail spam filter using novel TFDCR feature selection with dynamic feature update

Gopi Sanghani [a],[*], Ketan Kotecha [b]

[a] *Computer Engineering Department, Nirma University, Ahmedabad 382481, India*
[b] *Symbiosis Institute of Technology, Symbiosis International (Deemed University), Pune, India*

## ARTICLE INFO

*Article history:*
Received 4 September 2017
Revised 1 July 2018
Accepted 21 July 2018
Available online 24 July 2018

## ABSTRACT

Communication through e-mails remains to be highly formalized, conventional and indispensable method for the exchange of information over the Internet. An ever-increasing ratio and adversary nature of spam e-mails have posed a great many challenges such as uneven class distribution, unequal error cost, frequent change of content and personalized context-sensitive discrimination. In this research, we propose a novel and distinctive approach to develop an incremental personalized e-mail spam filter. The proposed work is described using three significant contributions. First, we applied a novel term frequency difference and category ratio based feature selection function TFDCR to select the most discriminating features irrespective of the number of samples in each class. Second, an incremental learning model is used which enables the classifier to update the discriminant function dynamically. Third, a heuristic function called *selectionRankWeight* is introduced to upgrade the existing feature set that determines new features carrying strong discriminating ability from an incoming set of e-mails. Three public e-mail datasets possessing different characteristics are used to evaluate the filter performance. Experiments are conducted to compare the feature selection efficiency of TFDCR and to observe the filter performance under both the batch and the incremental learning mode. The results demonstrate the superiority of TFDCR as the most effective feature selection function. The incremental learning model incorporating dynamic feature update function overcomes the problem of drifting concepts. The proposed filter validates its efficiency and feasibility by substantially improving the classification accuracy and reducing the false positive error of misclassifying legitimate e-mail as spam.

© 2018 Elsevier Ltd. All rights reserved.

## 1. Introduction

The incredible growth of the Internet has remarkably increased the number of e-mail users worldwide. Business and personal communication through e-mails remains to be highly formalized, conventional and indispensable method for the exchange of information even after increasing use of social networking applications. However, the inevitable downside of it is a continuously growing ratio of unwanted and useless e-mails called spam e-mails. Though e-mails are considered to be the most reliable medium, a massive number of unsolicited e-mails are delivered to the Internet users every day without any personal or commercial level of interest. Due to the availability of spamming tools & software and fastest & easiest delivery mode, spam e-mails have turned out to be the great nuisance for end users. An interruption caused by the continuous stream of spam e-mails results in wastage of human efforts and network bandwidth. Beginning with some usual advertising e-mails, over the time, around 50 to 80 percent worldwide e-mail traffic is now generated by spam e-mails. The adverse effect caused by spam e-mails has resulted in the economic loss of billions of dollars annually (Rao & Reiley, 2012). The characteristics of spam e-mails change rapidly over time as spammers continuously invent new strategies to evade spam filters. With the advent of technology and intelligent approaches, spammers have targeted more malicious and criminal activities creating even more attractive traps for the Internet users.

E-mails classification has remained to be an ongoing area of research because the adversary characteristics and capricious behavior of spam e-mails pose a great many challenges. The behavioral characteristics of spam e-mails can be specified by three types, as, *occurring once only*, i.e., spam e-mails delivered for a fixed period then disappear, *continuously occurring* spam e-mails and spam e-mails *reappear after some interval*, i.e., recurring context – a special subtype of concept drift (Widmer & Kubat, 1996). The rate at which spam e-mails are sent is unpredictable and reasonably different than the legitimate e-mails. The uneven proportion of spam

* Corresponding author.
 *E-mail address:* gopisanghani@gmail.com (G. Sanghani).

and legitimate e-mail traffic causes the problem of skewed class distribution (Fawcett, 2003). On the other side, the e-mail spam definition is getting more influenced by the personalized context and choice. Individual e-mail receivers can have different discrimination boundaries and preferences for considering the content of e-mails as either spam or legitimate. Often these preferences change with time as they are likely to follow individual's profile, job profile, interest, etc. Conventionally, in a spam filter, a spam message is considered as a positive instance, and a legitimate message is a negative instance. Misclassifying a legitimate e-mail to be spam (i.e., a false positive error) is usually more harmful than misclassifying a spam e-mail to be legitimate (i.e., a false negative error). As a result, an unequal and uncertain error cost is involved in the e-mail classification problem. The adversary nature and the menace caused by spam e-mails are the primary causes for the criticality of the issue.

The changing content of spam e-mails and changing target concepts result in concept drift in e-mail classification problem (Gama, Zliobaite, Bifet, Pechenizkiy, & Bouchachia, 2014). The change in the data distribution, i.e., the changing content of e-mails also causes to vary the discriminating ability of features; as a result, a feature distribution shift occurs. Global spam filters trained on general mail corpus work well for common spam categories. To emphasize the importance of different e-mail categories several services are also made available to the user inbox. Generally, static filters assume the same distribution of training data and testing data. But to deal with the drift, learned hypothesis of the discriminant function should be updated dynamically. The need also arises for modeling appropriately the communication pattern of an individual user in the classifier.

Content-based classification of e-mails as spam or legitimate is one of the most prominent applications of text classification problems. Recent advances in the development of spam filter confirm the success of employing machine learning algorithms for e-mail classification problem. Different approaches, proved to achieve the efficient results, rely more on the content analysis of e-mail body. Every day about 200 billion e-mails are sent to valid e-mail addresses (E-mail Statistics Report, 2013-2017). The spam volume varies non-monotonically. The challenges discussed here prove an e-mail classification problem to be more demanding than the classical text classification problems.

In this paper, we propose an incremental personalized e-mail spam filter for the classification of e-mails as spam or legitimate. The significant contributions of our research are summarized as follows:

i. The performance of the content-based classifier is affected by the discriminating ability of the selected features. To address the issue of uneven class distribution, we propose a novel term frequency difference and category ratio based feature selection function, named as TFDCR, for generating features with strong discriminating ability from the training data.
ii. We used the incremental learning model using Support Vector Machines (SVM) so that the learned decision model is updated to adjust the modified distribution of data in the presence of drifting concepts.
iii. Due to a frequent change in the content of e-mails, the relevance of the representative features also varies over a period. We propose a novel *selectionRankWeight* heuristic function based on the feature's category ratio difference to identify new features from an incoming set of e-mails. The existing feature set is updated by including these newly selected features before activating incremental training of the classifier.

Our distinctive approach successfully develops an intelligent expert system for personalized e-mail spam filtering using the contributions mentioned above. The function TFDCR is used to generate the feature space that creates the class representation model and forms the knowledge base of an expert system. The SVM classification model serves to be an inference engine of the expert system constructs an optimal decision hyperplane in this feature space and separates the two classes spam and legitimate. The learned decision function predicts the class information for the incoming set of e-mails. The incremental learning model and dynamic feature update function enhance the filtering system to relearn from the modified distribution of data and to be adaptive to classify the incoming set of e-mails correctly.

The remainder of this paper is organized into following sections. The research work related to this study is reviewed in Section 2. Section 3 presents our proposed algorithm for feature selection function TFDCR and the development of classifier. It also discusses the complete e-mail spam filtering system using incremental learning model with the inclusion of dynamic feature update function. Section 4 gives datasets description and various performance measures used in this study. It also describes the analysis of experiments performed for the evaluation of the filter and discussion on the results achieved. Section 5 concludes the paper with some insightful details of improvements and extensions to be carried out in the future.

## 2. Related work

With the increase in the criticality of spam e-mail problem, many efficient spam filtering techniques are implemented in the past few years. Various automatic e-mail classification techniques are rule-based approaches, white and blacklists, collaborative spam filtering and content-based spam filtering. Rule-based spam detection uses pattern analysis, pattern selection, and score assignment. However, these rules need to be regularly updated; rule-based systems also tend to have a high false positive rate if the rules are not reformulated suitably. Depending on the sender's authenticity, an approach called black/whitelist was presented that relies on the number of IP addresses to determine whether an e-mail is a spam e-mail (Jung & Sit, 2004). The blacklist includes an e-mail server and an IP address of the sender. If the source of an e-mail appears in the blacklist, the e-mail is identified as spam. The approach suffers from the problem of updating and maintaining the list.

Collaborative spam filtering exploits the shared network characteristics to distribute information concerning spam messages and servers. The e-mail spam filter employed by an individual user and categorizes e-mails at client-side works as a personalized spam filter. Gray and Haahr (2004) proposed an approach that used the concept of personalized, collaborative spam filtering called CASSANDRA architecture. They presented a collaborative spam filtering which was detailed, along with a proof-of-concept, peer-to-peer, signature-based implementation. Shih, Chiang, and Lin (2008) proposed a collaborative spam filtering using a multi-agent learning architecture for improving the ability of individual spam detection. The authors showed that though the signature-based methods have several weaknesses, under the collaborative scheme the proposed framework helped to build clusters for users' preferences in anti-spam prevention.

Machine learning algorithms have been proved to achieve the efficient performance in spam categorization (Cormack, 2008). Different machine learning algorithms such as Support Vector Machines, Naïve Bayes algorithms, case-based reasoning systems, artificial neural networks, artificial immune systems, etc. have been successfully applied to the spam filtering domain. Filtron (Michelakis, Androutsopoulos, Paliouras, Sakkis, & Stamatopoulos, 2004), a personalized anti-spam filter, had been evaluated in real life scenario using four learning algorithms, namely Naive Bayes, Flexible Bayes, Support Vector Machines and LogitBoost. The evaluation of filter confirmed the prominent role of machine learning for e-mail spam filtering.

Content-based classification techniques aim to use the textual content of e-mails by analyzing subject, body, and attachments. Cheng and Li (2006) presented an approach that combined supervised and semi-supervised classifiers using SVM for personalized spam filtering. Authors modeled distribution shift of training and testing data as a variation of decision hyperplane and showed that the combined classifier achieved higher accuracy. Teng and Teng (2008) proposed the two-tier spam filter structure in which, e-mails identified as legitimate e-mails by the legitimate mail filter may pass, while the remaining e-mails were processed ordinarily by the spam filter. They modified *tfidf* algorithm, implemented an SVM-based filter and observed that the two-tier structure reached much lower FP rate and the same FN rate.

Chang, Yih, and Mccann (2008) designed a light-weight and highly scalable user model to address the issue of grey e-mail that could be easily combined with a traditional global spam filter. The model incorporated the user feedback on message labels and recognized up to 40% more spam from gray e-mail in the low false-positive region. Youn and McLeod (2009) proposed a spam e-mail filtering method that used the user profile ontology to create a blacklist of contacts and topic words. The filter implemented with *tfidf* feature selection algorithm and the C4.5 decision tree as a classifier performed better with personalized ontology than global ontology. Junejo and Karim (2013) proposed an automatic approach to build a statistical model of spam and non-spam words from the labeled training dataset. Their personalized spam filter was updated in two passes over the unlabeled individual user's inbox to handle the distribution shift. Their experimental results demonstrated the robustness and effectiveness of the filter as a global and personalized service side spam filter. The filter presented by Shams and Mercer (2013) used natural language attributes, the majority of them being connected to stylometry aspects of writing. Random Forest, Support Vector Machine, Naive Bayes, AdaBoost, and Bagging were used to generate the classifiers. Santos, Laorden, Sanz, and Bringas (2012) explored the use of semantics in spam filtering by representing e-mails as the enhanced topic-based vector space model to detect the internal semantics of spam messages. The proposed model represented the linguistic phenomena using a semantic ontology and applied four machine learning algorithms Bayesian networks, decision tree, k-nearest neighbor and support vector machine.

The adversarial characteristics of spam e-mails motivated the development of dynamically updated spam filters to handle the drifting concepts. Delany, Cunningham, Tsymbal, and Coyle (2005) presented a case-based system for anti-spam filtering called ECUE that learned dynamically to track concept drift. The system updated the case-base at the end of each day with cases that were misclassified by the system that day, with a periodic rebuilding of the case-base using the most recent cases to reselect features. Fdez-Riverola, Iglesias, Diaz, Mendez, and Corchado (2007) proposed two new techniques for tracking concept drift in their fully automated instance-based reasoning (IBR) system for spam labeling and filtering. The authors showed that the lazy learner algorithms could handle the concept drift by permitting the easy update of the model when a new type of spam appears. Hsiao and Chang (2008) developed incremental cluster-based classification method, called ICBC which clusters e-mails in each given class into several groups, and an equal number of features (keywords) were extracted from each group. The authors proved experimentally ICBC can effectively address the issues of skewed and changing class distributions, and the cost of re-training was reduced by incremental learning the model. Katakis, Tsoumakas, and Vlahavas (2010) presented the general framework for classifying data streams by exploiting stream clustering to build and update an ensemble of incremental classifiers dynamically. Yevseyeva, Basto-Fernandes, Ruano-OrdaS, and MeNdez (2013) compared two most widely used multi-objective evolutionary algorithms NSGA-II and SPEA2 with a single objective evolutionary algorithm, Grindstone4SPAM. The authors concluded NSGA-II revealed the most promising results among the three algorithms. Georgala, Kosmopoulos, and Paliouras (2014) proposed active learning approach using incremental clustering for spam filtering.

Content-based classification algorithms require selecting the most discriminating features from the training set. An extensive collection of text documents may contain a substantial number of features, all of which cannot be included for the classification task. Feature selection aims to identify only features with the higher discriminating ability to improve the performance of the classifier. Yang and Pedersen (1997) presented a comparative study of many classical feature selection functions for text categorization. Lee and Lee (2006) introduced a new information gain and divergence-based feature selection method for statistical machine learning-based text categorization without relying on more complex dependence models. In the presence of concept drift the distribution of data changes due to which the relevance of features also varies. This variation causes the existing classification model to be inconsistent with the discrimination of new data. Katakis, Tsoumakas, and Vlahavas (2006) presented the idea of dynamic feature space and incremental learning for textual data classification. To handle the concept drift issue in stream data, they proposed a framework with two components, an incremental feature ranking method, and an incremental learning algorithm that could consider a subset of the features during prediction. Their experiments showed that the proposed feature based learning approach offered better predictive accuracy compared to classical incremental learning. Pinheiro, Cavalcanti, and Ren (2015) suggested two methods *maximum features per document* and *maximum features per document – reduced* for feature selection in text categorization. Both algorithms determined the number of selected features in a data-driven way using a global-ranking feature evaluation function. Wang, Liu, Feng, and Zhu (2015) proposed an existing optimal document frequency-based feature selection method (ODFFS) and a predetermined threshold applied to select the most discriminative features. Uysal (2016) proposed an improved global feature selection scheme IGFSS which was an ensemble method combining the power of filter based global and one-sided local feature selection method. Agnihotri, Verma, and Tripathi (2017) addressed an issue related to the equal or variable number of features from each class.

## 3. Proposed algorithm

Discrimination of e-mails based on an unknown and identical distribution often does not reflect individual user preferences. The primary purpose of our research is to develop an e-mail spam filter that captures the characteristics of a particular user from the known distribution of samples. For the filter to be the most effective we have defined a new term frequency difference and category ratio based feature selection function TFDCR. Our filtering system is enhanced by the ability to modify the existing feature set and subsequently to update the classifier incrementally. In this section, we describe the method for TFDCR feature selection function and the development of an incremental personalized e-mail spam filtering system with the proposed function for dynamic feature update.

**Algorithm 1** Feature selection using the novel TFDCR function.

1. Extract all unique features from legitimate and spam e-mails from training dataset.
2. Compute spam term and document occurrences and legitimate term and document occurrence of each feature from the training set.
3. Apply feature selection function on each feature to find its discriminative weight ($dmw$) as follows:
$$feature_i^{dmw} = (|termFreq_i^s - termFreq_i^l|) * product(termCatRatio_i)$$
Where,
$$product(termCatRatio_i) = \begin{cases} (\frac{DocFreq_i^s}{N_S} * \frac{N_L}{DocFreq_i^l}) & if(\frac{DocFreq_i^s}{N_S} > \frac{DocFreq_i^l}{N_L}) \\ (\frac{DocFreq_i^l}{N_L} * \frac{N_S}{DocFreq_i^s}) & otherwise \end{cases}$$
4. Sort the features in descending order of discriminative weight.
5. Select top N features for generating final feature set.

## 3.1. Feature selection

The appropriate combination of preprocessing tasks provides a significant improvement on classifier's performance (Uysal & Gunal, 2014). The preprocessing phase consists of functions such as tokenization, stop word removal and stemming. The next essential phase is of feature extraction and selection from vector space model representation. Content-based classification is highly dependent on the discriminating power of the selected features. An excessive number of features considerably increase the training time and memory overhead. Too many features can also have a negative impact on the classification accuracy which in turn will adversely affect the performance of the classifier (Uysal, 2016).

The selected feature set creates a representation model for a specific class or a category. A feature with high occurrence in any one class implicates its importance for that class. The category ratio of a feature also plays a significant role in addressing the issue of uneven class distribution. This specification leads us to the conclusion that the features with a higher difference of occurrence in both the categories and the higher category ratio tend to have the strong discriminating ability. The occurrence difference and category ratio are used to derive the proposed feature selection function TFDCR. Algorithm 1 shows the method for TFDCR feature selection function. Table 1 gives the meaning of notations used in Algorithm 1. To determine the discriminative weight of feature $i$, we first compute its distribution in both the categories. The first term of discriminative weight formula gives the occurrence difference of feature $i$. The second term identifies feature's distribution within a particular class. Two components are used to generate the second term. The first component is a category ratio, which increases the weight of features with higher document frequency than those features having higher term frequency but they are present only in a few documents. The other component is a reverse category ratio, which is used to emphasize more on features with lower document frequency of another class. The computation of discriminative weight using the proposed function ensures an automatic selection of features representing both the categories irrespective of the number of examples in each class. Also, the formula emphasizes very well on inter-class distribution as well as an intra-class distribution of a feature.

**Table 1**
Meaning of notations used in the Algorithm 1.

| Notation | Meaning |
| --- | --- |
| $feature_i^{dmw}$ | The discriminative weight of feature $i$ |
| $termFreq_i^s$ | Total occurrences of feature $i$ in spam e-mails |
| $termFreq_i^l$ | Total occurrences of feature $i$ in legitimate e-mails |
| $DocFreq_i^s$ | Total document occurrences of feature $i$ in spam e-mails |
| $DocFreq_i^l$ | Total document occurrences of feature $i$ in legitimate e-mails |
| $N_L$ | Total number of legitimate e-mails |
| $N_S$ | Total number of spam e-mails |

## 3.2. Classification

Support vector machines (SVM) (Cortes & Vapnik, 1995) are efficient supervised machine learning algorithms also known as optimal margin classifiers. The proposed incremental personalized e-mail spam filtering system uses SVM for classification of e-mails. Drucker, Wu, and Vapnik (1999) initially applied SVM for spam categorization. Joachims (1998) showed that SVM achieved state-of-the-art performance on text classification problems. SVM is a discriminative classifier that maps input vectors into a feature space of higher dimension and constructs an optimized hyperplane for generalization using kernel trick. In a binary classification problem, a dataset X contains $n$ labeled example vectors $\{(x_1, y_1) \ldots (x_n, y_n)\}$, where $x_i$ is the input vector in the input space, with corresponding binary labels $y_i \in \{-1, 1\}$. Let $\phi(x_i)$ be the corresponding vectors in feature space, where $\phi(x_i)$ is the implicit kernel mapping and let $k(x_i, x_j) = \phi(x_i).\phi(x_j)$ be the kernel function, implying a dot product in the feature space. The optimization problem for a soft-margin SVM is,

$$\min_{w,b} \left\{ \frac{1}{2}||w||^2 + C \sum_i \xi_i \right\} \quad (1)$$

Subject to the constraints $y_i(w.x + b) = 1 - \xi_i$ and $\xi_i \geq 0$ where $w$ represents the normal vector of the separating hyperplane in feature space. The regularization parameter $C > 0$ controls the penalty for misclassification error. Eq. (1) is referred to as the primal equation. From that, the Lagrangian form of the dual problem is:

$$w(\alpha) = \max_\alpha \left\{ \sum_i \alpha_i - \frac{1}{2} \sum_{i,j} \alpha_i \alpha_j y_i y_j \, k(x_i, x_j) \right\} \quad (2)$$

Subject to $0 \leq \alpha_i \leq C$. This is a quadratic optimization problem that can be solved efficiently using algorithms such as Sequential Minimal Optimization (Platt, 1999). Many $\alpha_i$ go to zero during optimization and the remaining $x_i$ corresponding to those $\alpha_i > 0$ are called support vectors. Support vectors represent feature space and class boundaries in a very concise manner. If $l$ is the number of support vectors and $\alpha_i > 0$ for all $i$, with this formulation, the normal vector of the separating plane $w$ is calculated as:

$$w = \sum_{i=1}^{l} \alpha_i y_i x_i \quad (3)$$

The classification $f(x)$ for a new sample vector $x$ can be determined by computing the kernel function of $x$ with every support vector:

$$f(x) = sign\left( \sum_{i=1}^{l} \alpha_i y_i . \, k(x, x_i) + b \right) \quad (4)$$

Algorithm 2 describes the proposed system for incremental personalized e-mail spam filtering with dynamic feature update function. The system comprises three passes. The first of which

**Algorithm 2** Incremental personalized e-mail spam filter with dynamic feature update.

Input:
1. Training Set $Trem^0 = \{Em_s\} \cup \{Em_l\}$
2. $\rho \leftarrow$ threshold value for Accuracy
3. $Trem^0 \leftarrow$ n training e-mails with labels
4. $Em_s \leftarrow$ Set of Spam e-mails
5. $Em_l \leftarrow$ Set of Legitimate e-mails

Step1 Pre-processing of training & testing sets $Trem^0$ & $Tsem$
- Tokenization
- Stop word removal
- Stemming

Step 2 Feature Selection using TFDCR described in Algorithm 1

Step 3 SVM Conventional batch Training using SMO Algorithm (Pass I)
  Output: Support Vector Set $\alpha_i = \{\alpha_k \mid k = 1 \text{ to } l\}$
  $i = 0$

Step 4 Testing Phase (Pass II)
  Input: Testing Sets $Tsem = \{Ts^1, Ts^2, ..., Ts^m\}$
  // Testing instances contain set of unlabeled spam and legitimate e-mails
  Repeat
   i  Classify testing instances $Ts^1, Ts^2 ...$ from $Tsem$
   ii Add misclassified e-mails to set $Mcm_i$
  until either accuracy $\leq \rho$ or FPR increases
  $i = i + 1$

Step 5 Incremental SVM Training with updated feature set (Pass III)
  Input:
   i  Resulting Support Vector Set $\alpha_{i-1}$
   ii Re-training set $Rtrem_i = Mcm_{i-1} \cup \alpha_{i-1} \cup Ts_k$,
  where $Ts_k$ is the testing instance for which accuracy / FPR constraint is violated.
  5.1 Update Features as follows,
   i  $FS_i = FS_{i-1}$.
   ii Generate a new set of Features of the same dimension as $FS_{i-1}$ from $Rtrem_i$ and find a subset of distinct features from it. Call it $DNFS_i$
   iii For each feature $f_j$ in $DNFS_i$ find its *selectionRankWeight* (sRW) as,

$$sRW(f_j) = \left|\left(\frac{DocFreq_j^s}{N_S} - \frac{DocFreq_j^l}{N_L}\right)\right| * \left|\left(\frac{termFreq_j^s - termFreq_j^l}{termFreq_j^s + termFreq_j^l}\right)\right|$$

   iv Replace old features from $FS_i$ as
     if $sRW(f_j) > $ Average $sRW(DNFS_i)$ then
      $FS_i = FS_i \cup \{f_j\}$ &

  $FS_i = FS_i - - \{f_q \mid f_q$ has lowest $dmw\}$
  5.2 Retrain SVM on $Rtrem_i$
   Output Support Vector Set $\alpha_i = \{\alpha_k \mid k = 1 \text{ to } r\}$
   Repeat step 4 & 5 with $k = k + 1$

is a conventional batch learning using SVM. During pass I of Algorithm 2, the training set that consists of labeled spam and legitimate e-mails is given as input. Pass I creates the classification model $F(x)$. Once the classifier is built, a testing phase takes place in pass II. During the testing phase, a series of unlabeled e-mails are submitted to identify their true labels. The testing phase is continued until the validation criteria are violated. We consider two parameters, accuracy and false positive rate (FPR) as validation criteria. Either the classification accuracy decreases than the predefined threshold or the false positive rate increases indicates that the current model is inappropriate for the discrimination of incoming set of e-mail. In the e-mails classification, the misclassification of legitimate as spam (false positive) is more harmful than misclassification of spam as legitimate (false negative). Therefore, we set these two parameters as validation criteria for the classification model. Violation of any one of the parameters will cause the feature set to be updated and activate incremental learning of SVM, further explained in the subsequent subsection.

### 3.3. Incremental learning

One of the significant parameters of success of machine learning algorithms in classification problems is the adaptability towards incremental learning process. Incremental learning is a machine learning paradigm where the current model relearns whenever new example(s) appear over a period and contribute some different knowledge to the existing hypothesis. Incremental learning possesses the ability to include additional training data when it becomes available and to adjust the current decision model. Incremental SVM (Laskov, Gehl, Kruger, & Muller, 2006; Shilton, Palaniswami, Ralph, & Tsoi, 2005; Syed, Liu, & Sung, 1999) learning involves re-training a support vector machine after adding a small number of additional training vectors to the existing set of support vectors. In conventional batch learning mode, the learned model exhibits the knowledge derived from the nature and characteristics of training data. As long as the statistical distribution does not change in test data, the decision model performs well. The model performance is degraded when the underlying data distribution varies. The content of legitimate and spam e-mails change over a time which causes the phenomena of concept drift resulting in either the feature distribution shift or the change of class label or both. So, e-mail classification task requires a classifier to be trained and updated incrementally to handle the change of data distribution and at the same time to hold the previously acquired knowledge.

In SVM model, a set of support vectors precisely provides a representation of the training examples for the given classification task. Re-training using new examples and a set of support vectors allows the decision model to relearn the modified distribution of data. If the statistical distribution of new batch and the whole dataset does not differ much then the resulting decision function can be roughly similar. But in the presence of concept drift, the statistical distribution of a new batch of samples varies; incremental learning ensures that the concept learned in the previous step incorporates new definition such that, the modified hypothesis preserves and represents the decision model precisely. It is entirely possible that the separating hyperplane would be optimized in such a way that a new set of support vectors will be generated after re-training. Due to a change in the nature of an incoming set of e-mails or shift in the target information, the class boundary differs than that of the previous training phase. Fig. 1 shows the schematic representation of proposed incremental personalized e-mail spam filtering system.

Our incremental learning framework incorporates the function to update the feature set dynamically to handle the issue of feature shift distribution. Eq. (5) shows the formula for the proposed *selectionRankWeight* heuristic function. The heuristic function calculates the product of category ratio difference and normalized term difference for identifying new features with strong discriminating ability from the re-training set. We define *selectionRankWeight* as,

$$selectionRankWeight(f_j) = \left|\left(\frac{DocFreq_j^s}{N_S} - \frac{DocFreq_j^l}{N_L}\right)\right| * \left|\left(\frac{termFreq_j^s - termFreq_j^l}{termFreq_j^s + termFreq_j^l}\right)\right| \quad (5)$$

Only features with higher *selectionRankWeight* than the average weight of newly found distinct feature set are included to generate the updated feature set as depicted in step 5.1 of Algorithm 2. An equal number of old features with the lowest discriminative weight is removed to maintain the same feature dimensionality. The updated feature set is used to represent the re-training set of e-mails. The incremental training is activated to update the classification model, and a modified decision function $F'(x)$ is generated. The subsequent testing phase will be continued using this modified decision model. Likewise, the classifier is incrementally

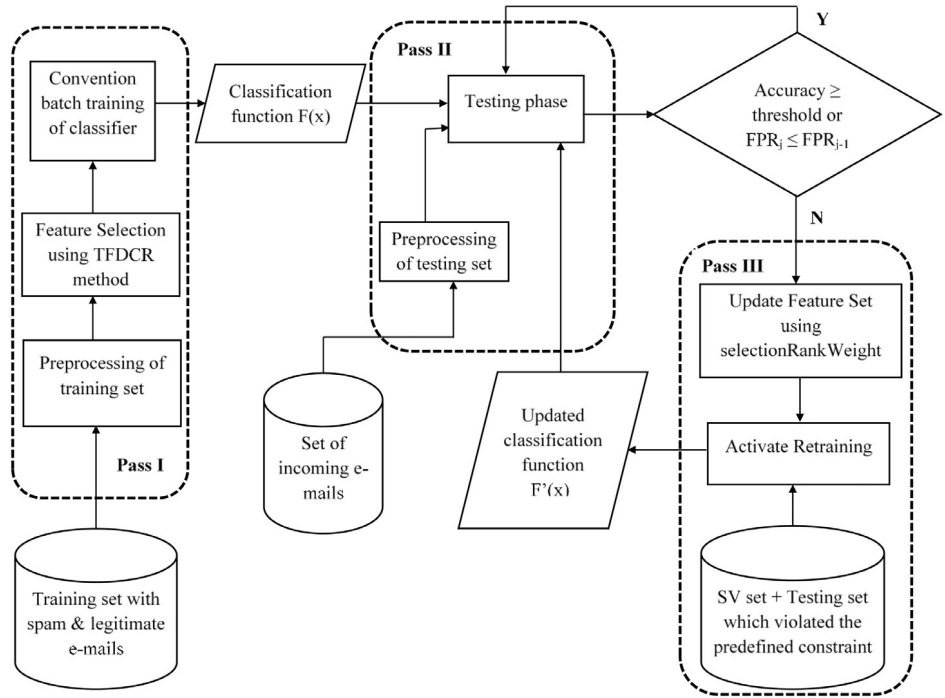

**Fig. 1.** Incremental personalized e-mail spam filter.

updated as and when there is a violation of any of the validation criteria.

## 4. Experiments and results

The proposed research work is validated with the help of in-depth experiments with three different public datasets. Two different types of experiments are performed to illustrate the efficiency of TFDCR feature selection function and the applicability of incremental personalized e-mail spam filter. Datasets, experimental methodology, and results are analyzed and discussed in detail in subsequent subsections.

### 4.1. Datasets

Three benchmark datasets with different characteristics are used for performing experiments and measuring success parameters. The first dataset is a public ENRON dataset (Enron Spam Data sets, 2006). This dataset contains pre-processed e-mail messages with the removal of attachments. The dataset includes six e-mail folders of individuals, farmer-d, Kaminski-v, kitchen-l, Williams-w3, beck-s, and lokay-m, named as Enron1 to Enron6. In three of the folders, the ratio of legitimate to spam e-mails is approximately 3:1, while in the other three the ratio is inverted to 1:3. The total number of messages in each dataset is between five and six thousand. The six folders of Enron dataset contain different types of e-mails received by real users; help to evaluate the performance of personalized e-mail spam filter. Moreover, the e-mails are numbered as per the order of arrival, which provides a suitable framework to apply incremental learning. The second dataset used for filter evaluation is ECML. The ECML task-A and task-B datasets were made publically available during 2006 ECML-PKDD Discovery Challenge (Discovery Challenge, 2006). Token id along with its count is used to represent e-mails in this dataset rather than actual content. Both ECML task-A and task-B training and test datasets follow the similarity in the proportion of spam and legitimate e-mails only. The source of data in training and testing follow a different distribution. Experiments on ECML datasets are conducted to analyze the personalized e-mail spam filter performance in the presence of distribution shift. The third dataset considered is PU dataset which contains four folders PU1, PU2, PU3 and PUA that contain e-mails received by particular users (Androutsopoulos, Koutsias, Chandrinos, & Spyropoulos, 2000). The arrival order of the e-mails is not preserved in this dataset. The proportion of spam and legitimate e-mails along with the characteristics of each dataset are given in Table 2.

### 4.2. Performance measures

The filter is evaluated with well-known performance measures used in classification. We measure accuracy, false positive rate and false negative rate defined in Eqs. (6)–(8).

$$Accuracy = (|n_{l \to l}| + |n_{s \to s}|)/(N_L + N_S) \quad (6)$$

$$false\ positive\ rate\ (FPR) = |n_{l \to s}|/N_L \quad (7)$$

$$false\ negative\ rate\ (FNR) = |n_{s \to l}|/N_S \quad (8)$$

Where, $N_S$ and $N_L$ are total spam and legitimate e-mails respectively, $n_{l \to l}$ is the number of legitimate e-mails correctly classified, i.e., True Negatives (TN), $n_{s \to s}$ is the number of correctly classified spam e-mails, i.e., True Positives (TP), $n_{l \to s}$ is the number of misclassified legitimate e-mails, i.e., False Positives (FP) and $n_{s \to l}$ is the number of spam e-mails not correctly classified, i.e., False Negatives (FN). The other two success measures employed are micro-F1 and macro-F1 defined using Eqs. (9) and (10) respectively.

$$Micro - F1 = (2 \times P \times R)/(P + R) \quad (9)$$

$$Macro - F1 = \sum_{k=1}^{C} F_k/C \text{ where } F_k = (2 \times P \times R)/(P + R) \quad (10)$$

Where P and R denote precision and recall measures. In machine learning Matthews correlation coefficient (MCC) (Matthews, 1975) is used as a measure of the quality of binary classification, MCC equals to +1 indicates a perfect prediction, 0 an average random prediction and −1 an inverse prediction. MCC is determined as follows:

$$MCC = \frac{(|TP| \cdot |TN|) - (|FP| \cdot |FN|)}{\sqrt{(|TP| + |FP|) \cdot (|TP| + |FN|) \cdot (|TN| + |FP|) \cdot (|TN| + |FN|)}} \quad (11)$$

**Table 2**
Description of e-mail datasets.

| Datasets | Total E-mails | | Characteristics |
|---|---|---|---|
| | Spam | Legitimate | |
| Enron1 | 1500 | 3672 | • Different proportion of spam and legitimate e-mails |
| Enron2 | 1496 | 4361 | • E-mails sorted in date order |
| Enron3 | 1500 | 4012 | • Legitimate e-mails are personalized |
| Enron4 | 4500 | 1500 | |
| Enron5 | 3675 | 1500 | |
| Enron6 | 4500 | 1500 | |
| ECMLA U00 to U02 | 2000 (Training) 1250 (Testing) | 2000 (Training) 1250 (Testing) | • Training and test data follow the completely different distribution |
| ECMLB U00 to U14 | 50 (Training) 200 (Testing) | 50 (Training) 200 (Testing) | • Test data folders are personalized |
| PU1 | 481 | 618 | • E-mails are not sorted as per arrival time |
| PU2 | 142 | 579 | • Different folders are personalized |
| PU3 | 1826 | 2313 | |
| PUA | 571 | 571 | |

**Table 3**
The classification results in experiment 1.

| Dataset | Accuracy (%) | | | | | | MCC | | | | | |
|---|---|---|---|---|---|---|---|---|---|---|---|---|
| | TFDCR | IG | CHI | GINI | IGR | CFS | TFDCR | IG | CHI | GINI | IGR | CFS |
| ENRON1 | 93.82 | **94.18** | 93.69 | 93.28 | 88.02 | 93.82 | 0.85 | 0.86 | 0.85 | 0.84 | 0.73 | 0.85 |
| ENRON2 | **89.30** | 87.90 | 87.92 | 87.97 | 86.84 | 87.99 | 0.70 | 0.65 | 0.66 | 0.66 | 0.63 | 0.66 |
| ENRON3 | 93.08 | 93.83 | **95.17** | **95.17** | 88.57 | **95.17** | 0.82 | 0.84 | 0.88 | 0.88 | 0.73 | 0.88 |
| ENRON4 | **98.73** | 96.49 | 95.71 | 95.68 | 91.2 | 95.75 | 0.96 | 0.91 | 0.88 | 0.87 | 0.76 | 0.88 |
| ENRON5 | 92.23 | 96.14 | **96.26** | 96.08 | 93.44 | 96.23 | 0.82 | 0.91 | 0.91 | 0.90 | 0.84 | 0.91 |
| ENRON6 | **95.98** | 91.580 | 90.66 | 90.64 | 90.04 | 90.55 | 0.88 | 0.78 | 0.76 | 0.76 | 0.74 | 0.76 |
| ECMLAU00 | **62.00** | 60.60 | 56.56 | 56.36 | 52.12 | 53.32 | 0.24 | 0.22 | 0.15 | 0.14 | 0.06 | 0.09 |
| ECMLAU01 | **64.52** | 60.56 | 63.28 | 62.96 | 57.12 | 55.68 | 0.29 | 0.24 | 0.28 | 0.28 | 0.17 | 0.13 |
| ECMLAU02 | 71.16 | 67.40 | **72.60** | 72.48 | 64.32 | 61.96 | 0.42 | 0.43 | 0.50 | 0.50 | 0.37 | 0.32 |
| ECMLBU00 | **64.75** | 57.00 | 64.25 | 66.00 | 61.25 | 50.00 | 0.30 | 0.17 | 0.29 | 0.32 | 0.23 | 0.06 |
| ECMLBU01 | **70.25** | 59.75 | 66.00 | 65.75 | 54.75 | 50.75 | 0.41 | 0.28 | 0.33 | 0.32 | 0.10 | 0.02 |
| ECMLBU02 | **84.75** | 57.50 | 64.00 | 65.25 | 65.00 | 67.75 | 0.70 | 0.21 | 0.28 | 0.31 | 0.30 | 0.36 |
| ECMLBU03 | **84.75** | 69.50 | 50.75 | 68.25 | 58.50 | 65.00 | 0.70 | 0.43 | 0.02 | 0.37 | 0.19 | 0.32 |
| ECMLBU04 | **77.50** | 75.00 | 66.00 | 63.50 | 71.75 | 70.75 | 0.55 | 0.52 | 0.32 | 0.27 | 0.48 | 0.44 |
| ECMLBU05 | **81.50** | 53.00 | 63.50 | 40.00 | 68.50 | 70.25 | 0.64 | 0.10 | 0.27 | 0.20 | 0.37 | 0.41 |
| ECMLBU06 | **71.00** | 64.25 | 58.25 | 56.75 | 53.25 | 51.75 | 0.43 | 0.34 | 0.17 | 0.14 | 0.07 | 0.04 |
| ECMLBU07 | **74.75** | 62.25 | 53.00 | 51.25 | 51.00 | 61.00 | 0.52 | 0.31 | 0.06 | 0.03 | 0.04 | 0.22 |
| ECMLBU08 | **74.50** | 47.25 | 53.50 | 60.00 | 55.75 | 61.00 | 0.49 | 0.14 | 0.44 | 0.02 | 0.12 | 0.07 |
| ECMLBU09 | **80.50** | 54.25 | 68.00 | 50.75 | 62.25 | 53.50 | 0.61 | 0.13 | 0.37 | 0.45 | 0.25 | 0.06 |
| ECMLBU10 | **80.00** | 62.00 | 57.00 | 72.50 | 60.75 | 52.75 | 0.60 | 0.25 | 0.14 | 0.16 | 0.23 | 0.29 |
| ECMLBU11 | **74.50** | 57.75 | 61.00 | 58.00 | 63.50 | 64.25 | 0.49 | 0.16 | 0.22 | 0.22 | 0.27 | 0.27 |
| ECMLBU12 | **69.00** | 66.50 | 48.75 | 60.75 | 53.50 | 63.25 | 0.38 | 0.33 | 0.03 | 0.05 | 0.07 | 0.14 |
| ECMLBU13 | **78.00** | 60.50 | 68.00 | 47.75 | 67.75 | 56.75 | 0.56 | 0.22 | 0.36 | 0.35 | 0.37 | 0.04 |
| ECMLBU14 | **76.75** | 68.25 | 47.00 | 67.50 | 50.75 | 66.25 | 0.54 | 0.36 | 0.06 | 0.07 | 0.02 | 0.33 |
| PU1 | **96.75** | 95.81 | 96.21 | 94.45 | 93.37 | 95.26 | 0.93 | 0.91 | 0.92 | 0.89 | 0.87 | 0.90 |
| PU2 | 94.27 | **95.12** | 93.42 | 92.99 | 91.93 | 92.14 | 0.81 | 0.84 | 0.78 | 0.77 | 0.74 | 0.80 |
| PU3 | 95.50 | 95.75 | **96.15** | 95.34 | 95.14 | 95.46 | 0.91 | 0.92 | 0.92 | 0.91 | 0.90 | 0.92 |
| PUA | **95.20** | 94.10 | 90.77 | 93.17 | 91.70 | 94.28 | 0.90 | 0.88 | 0.82 | 0.86 | 0.84 | 0.89 |

### 4.3. Analysis of experiments

We conducted two different types of experiments to estimate the performance of the proposed research work. The first experiment is performed to analyze how effectively TFDCR feature selection function identifies the most discriminating features from the given training data. We used five basic and well-known feature selection functions Information Gain (IG), Chi-square (Yang & Pedersen, 1997), Gini index (Breiman, Friedman, Stone, & Olshen, 1984), Information Gain Ratio (IGR) (Karegowda, Manjunath, & Jayaram, 2010) and correlation feature selection (CFS) (Hall, 2000) to compare the results achieved through proposed TFDCR. The first experiment is conducted using a conventional batch training of classifier followed by a testing phase. Table 3 reports the accuracy of classification and MCC achieved through all feature selection functions for all the datasets. Three of six Enron datasets we achieved the highest accuracy, except two cases of ECML data-sets our results are superior as compared to other functions. And two of four datasets of PU we get higher accuracy with TFDCR feature selection function.

Table 4 gives the comparison of FPR and FNR for all the datasets considered. The FPR and FNR are averaged over all the folders of the individual dataset. The results using TFDCR function show that the lowest FPR is achieved in all the cases. Table 5 shows two other success measures Micro-F1 and Macro F1. Figs. 2–4 show the comparison graph for FPR and FNR achieved through all the feature selection functions for Enron, ECML and PU datasets respectively. An occurrence of false positive, i.e., legitimate e-mails incorrectly classified as spam is a substantial performance degrader for

**Table 4**
Average FPR and FNR achieved in experiment 1.

| Dataset | Average FPR | | | | | | Average FNR | | | | | |
|---|---|---|---|---|---|---|---|---|---|---|---|---|
| | TFDCR | IG | CHI | GINI | IGR | CFS | TFDCR | IG | CHI | GINI | IGR | CFS |
| ENRON | 0.049 | 0.073 | 0.069 | 0.07 | 0.097 | 0.069 | 0.119 | 0.115 | 0.114 | 0.113 | 0.144 | 0.498 |
| ECML A | 0.373 | 0.581 | 0.564 | 0.57 | 0.743 | 0.74 | 0.308 | 0.161 | 0.152 | 0.151 | 0.099 | 0.503 |
| ECML B | 0.198 | 0.227 | 0.445 | 0.478 | 0.553 | 0.47 | 0.277 | 0.605 | 0.495 | 0.494 | 0.337 | 0.499 |
| PU | 0.034 | 0.045 | 0.045 | 0.054 | 0.067 | 0.046 | 0.099 | 0.072 | 0.103 | 0.092 | 0.103 | 0.529 |

**Table 5**
The classification results in experiment 1.

| Dataset | Micro-F1 | | | | | | Macro-F1 | | | | | |
|---|---|---|---|---|---|---|---|---|---|---|---|---|
| | TFDCR | IG | CHI | GINI | IGR | CFS | TFDCR | IG | CHI | GINI | IGR | CFS |
| ENRON1 | 89.54 | **90.35** | 89.48 | 88.90 | 81.25 | 89.53 | 92.58 | 93.15 | 92.54 | 92.11 | 86.23 | **92.58** |
| ENRON2 | **74.25** | 71.47 | 72.72 | 72.91 | 69.37 | 72.80 | **83.75** | 81.82 | 82.48 | 82.59 | 80.50 | 82.55 |
| ENRON3 | 87.13 | 88.38 | 90.91 | 90.93 | 80.57 | **90.93** | 91.20 | 92.08 | 93.81 | 93.82 | 86.24 | **93.82** |
| ENRON4 | **99.19** | 97.96 | 97.25 | 97.24 | 94.21 | 97.28 | **98.15** | 95.13 | 93.75 | 93.72 | 87.96 | 93.81 |
| ENRON5 | 94.38 | **97.43** | 97.40 | 97.28 | 95.35 | 97.38 | 90.90 | 95.39 | **95.37** | 95.15 | 92.13 | 95.33 |
| ENRON6 | **97.48** | 94.39 | 93.79 | 93.77 | 93.36 | 93.71 | **93.80** | 88.56 | 87.50 | 87.48 | 86.74 | 87.37 |
| ECMLAU00 | 64.47 | 64.63 | 64.37 | 64.31 | 65.13 | **65.09** | **61.82** | 60.08 | 54.37 | 54.08 | 44.37 | 47.33 |
| ECMLAU01 | 65.61 | 67.88 | **68.84** | 68.67 | 66.25 | 65.00 | **64.48** | 58.40 | 62.07 | 61.69 | 53.74 | 52.29 |
| ECMLAU02 | 71.03 | 74.71 | **77.55** | 77.50 | 73.00 | 71.33 | **71.16** | 64.42 | 70.20 | 71.04 | 60.20 | 57.41 |
| ECMLBU00 | **62.60** | 44.55 | 29.97 | 31.96 | 42.89 | 51.66 | **64.63** | 41.32 | 35.31 | 33.94 | 38.43 | 52.73 |
| ECMLBU01 | **67.75** | 38.46 | 26.67 | 27.95 | 46.97 | 50.38 | **70.07** | 35.90 | 33.33 | 33.74 | 45.19 | 50.75 |
| ECMLBU02 | **84.32** | 64.11 | 60.44 | 62.13 | 65.69 | 67.67 | **84.74** | 51.37 | 63.71 | 65.01 | 64.99 | 67.75 |
| ECMLBU03 | **84.71** | 60.90 | 53.21 | 71.20 | 65.56 | 69.70 | **84.75** | 67.95 | 50.61 | 67.91 | 56.68 | 64.14 |
| ECMLBU04 | **78.16** | 70.76 | 68.52 | 60.96 | 76.70 | 74.62 | **77.48** | 74.46 | 65.78 | 63.35 | 70.41 | 70.05 |
| ECMLBU05 | **81.42** | 61.01 | 60.11 | 44.70 | 69.42 | 70.90 | **81.74** | 43.78 | 63.23 | 39.56 | 68.47 | 70.23 |
| ECMLBU06 | **68.13** | 51.19 | 45.18 | 54.83 | 56.44 | 57.40 | **70.76** | 61.49 | 41.52 | 56.67 | 43.98 | 50.89 |
| ECMLBU07 | **70.38** | 45.09 | 45.98 | 55.38 | 47.47 | 60.61 | **74.19** | 58.16 | 52.19 | 50.83 | 47.99 | 61.00 |
| ECMLBU08 | 72.87 | 62.76 | **73.85** | 40.82 | 56.08 | 54.63 | **74.41** | 33.29 | 71.27 | 48.20 | 55.75 | 53.47 |
| ECMLBU09 | **80.40** | 25.91 | 24.86 | 23.68 | 41.96 | 49.40 | **80.50** | 46.41 | 31.38 | 27.32 | 37.42 | 47.15 |
| ECMLBU10 | **78.84** | 55.81 | 56.57 | 58.42 | 65.80 | 67.43 | **79.94** | 61.24 | 57.00 | 58.00 | 59.88 | 63.91 |
| ECMLBU11 | **72.58** | 50.44 | 57.61 | 59.22 | 66.51 | 65.73 | **74.37** | 56.81 | 60.75 | 60.69 | 63.20 | 63.06 |
| ECMLBU12 | **68.04** | 65.99 | 47.84 | 48.40 | 59.21 | 61.81 | **68.97** | 66.49 | 48.73 | 47.74 | 52.57 | 55.98 |
| ECMLBU13 | **78.54** | 58.36 | 65.96 | 66.84 | 71.40 | 58.62 | **77.99** | 60.62 | 67.88 | 67.49 | 67.22 | 50.74 |
| ECMLBU14 | **76.57** | 67.01 | 51.38 | 52.86 | 60.52 | 67.47 | **76.75** | 67.97 | 46.57 | 45.51 | 47.54 | 66.20 |
| PU1 | **96.20** | 95.10 | 95.54 | 93.58 | 92.08 | 94.42 | **96.68** | 95.72 | 96.12 | 94.35 | 93.19 | 95.15 |
| PU2 | 83.02 | **87.01** | 82.08 | 81.36 | 78.65 | 83.91 | 89.79 | **92.00** | 89.02 | 88.52 | 86.84 | 90.13 |
| PU3 | 95.02 | 95.28 | 95.73 | 94.87 | 94.54 | 95.52 | 95.52 | 95.76 | **96.16** | 95.36 | 95.11 | 95.96 |
| PUA | **95.29** | 94.24 | 90.77 | 93.26 | 92.20 | 94.45 | **95.20** | 94.09 | 90.77 | 93.17 | 91.66 | 94.27 |

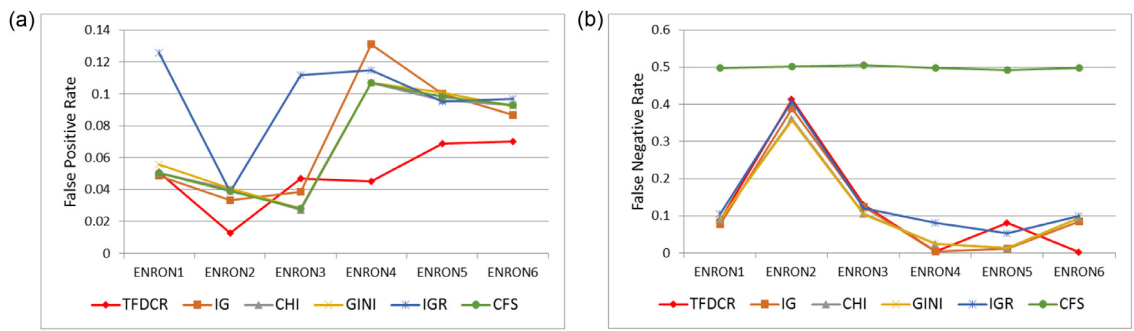

**Fig. 2.** Results of Enron dataset (a) false positive rate (b) false negative rate.

e-mail spam filters. Figs. 5(a)–(c) represent ROC curves in all three datasets for the comparison of true positive vs. false positive.

In the second experiment, we integrate an incremental learning model using SVM with an updated feature set to relearn the modified distribution of data. In this experiment, we use TFDCR feature selection in both the conventional batch learning and the incremental learning models. Features are updated using a *selectionRankWeight* heuristic function described in Eq. (5). The function finds new features from the re-training data with strong discriminating ability. Over an entire span of time, both the traditional features and latest features are essential for efficient filtering. Updating feature set helps to improve classification results particularly in the presence of drift. The experimental results confirm the applicability of the incremental learning model to modify the classifier to preserve better classification performance. A substantial reduction is achieved in FPR when the classifier is incrementally updated, even in the case when the distribution of data is different in training and testing sets as depicted in Table 6. Table 6 gives the comparison of both average FPR and FNR in incremental as well as the conventional batch model. Results also validate the incremental learning ability of SVM to tackle concept drift and improvement of prediction accuracy. Table 7 shows the comparison of classification results achieved in both the conventional batch model and incremental learning model for all three datasets. Incremental learning enables the classifier to be dynamically updated which results in better classification accuracy and significant decrement in FPR. Especially in case of ECML datasets where batch model accuracy is lower due to different test and training data distribution, the incremental learning achieves significantly improved results. Figs. 6–8 show FPR and FNR achieved for all three datasets. Fig. 9(a)–(c) represent ROC curves produced in all three datasets for both batch and incremental model.

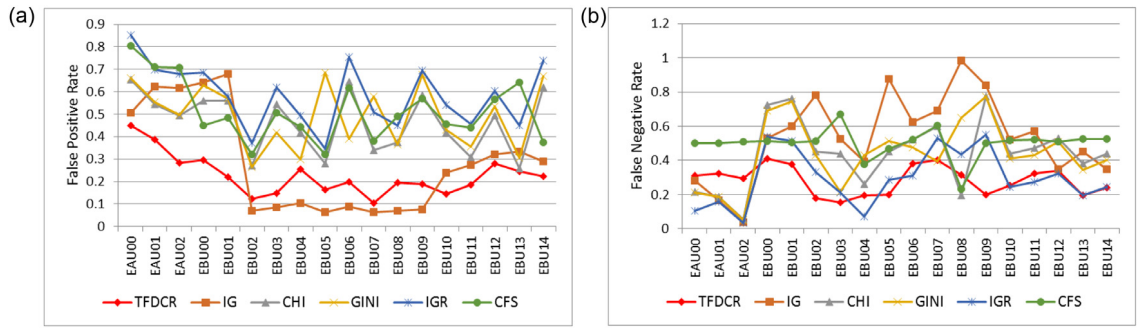

**Fig. 3.** Results of ECML dataset (a) false positive rate (b) false negative rate.

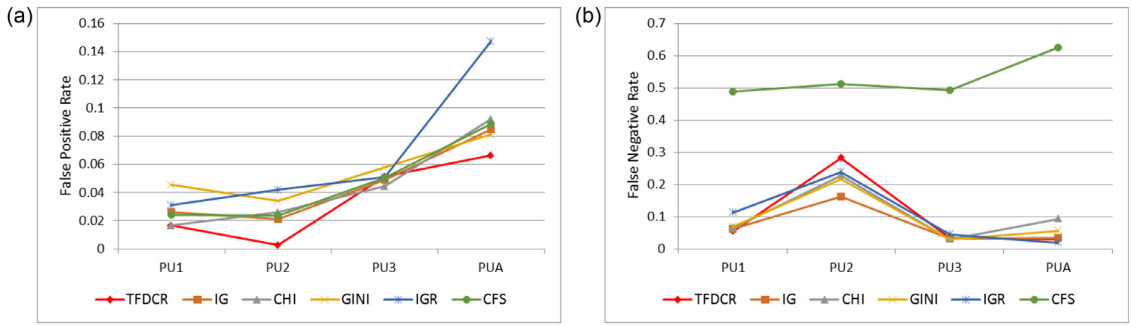

**Fig. 4.** Results of PU dataset (a) false positive rate (b) false negative rate.

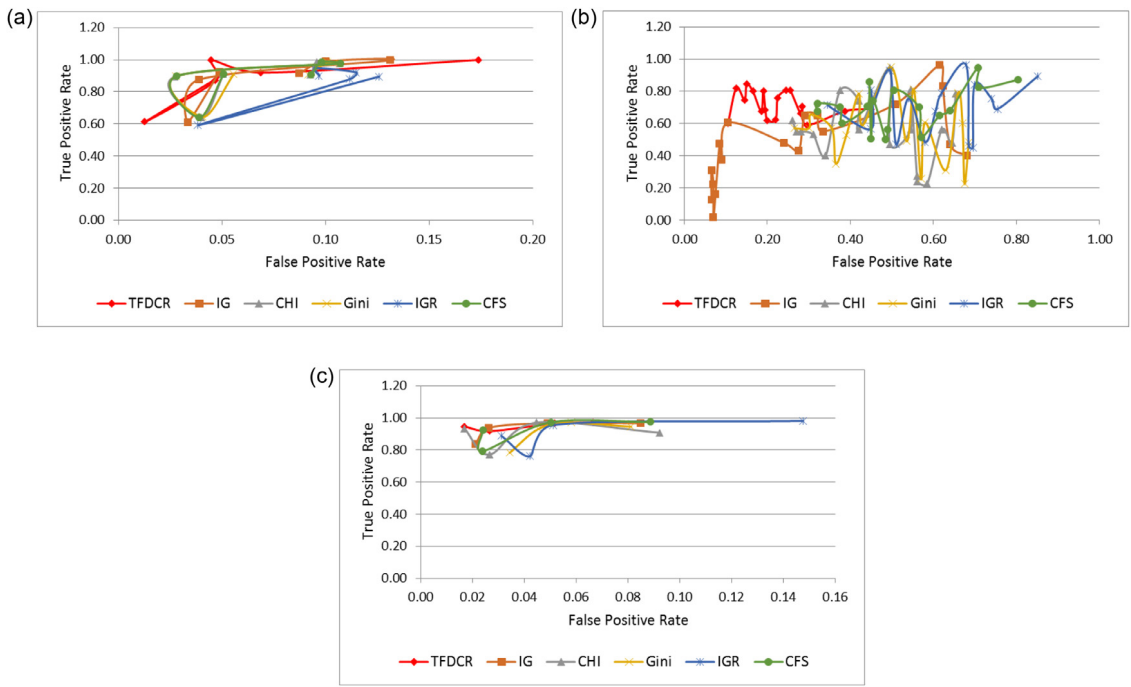

**Fig. 5.** ROC curve (a) Enron dataset (b) ECML dataset. (c) ROC curve PU dataset.

**Table 6**
Average FPR and FNR achieved in experiment 2.

| Dataset | Average FPR | | Average FNR | |
|---|---|---|---|---|
| | Incremental filter with TFDCR | Batch filter with TFDCR | Incremental filter with TFDCR | Batch filter with TFDCR |
| ENRON | 0.029 | 0.049 | 0.039 | 0.119 |
| ECML A | 0.005 | 0.373 | 0.026 | 0.308 |
| ECML B | 0.025 | 0.198 | 0.059 | 0.277 |
| PU | 0.023 | 0.034 | 0.065 | 0.099 |

**Table 7**
The classification results in experiment 2.

| Dataset | Accuracy (%) | | MCC | | Micro-F1 | | Macro-F1 | |
|---|---|---|---|---|---|---|---|---|
| | **Incremental SVM Training** | Batch SVM Training | **Incremental SVM Training** | Batch SVM Training | **Incremental SVM Training** | Batch SVM Training | **Incremental SVM Training** | Batch SVM Training |
| ENRON1 | 96.69 | 93.82 | 0.92 | 0.85 | 94.31 | 89.54 | 95.99 | 92.58 |
| ENRON2 | 97.38 | 89.30 | 0.93 | 0.70 | 94.68 | 74.25 | 96.47 | 83.75 |
| ENRON3 | 96.74 | 93.08 | 0.92 | 0.82 | 93.95 | 87.13 | 95.86 | 91.20 |
| ENRON4 | 99.20 | 98.73 | 0.98 | 0.96 | 99.49 | 99.19 | 98.83 | 98.15 |
| ENRON5 | 97.25 | 92.23 | 0.93 | 0.82 | 98.05 | 94.38 | 96.69 | 90.90 |
| ENRON6 | 97.63 | 95.98 | 0.93 | 0.88 | 98.47 | 97.48 | 96.60 | 93.80 |
| ECMLAU00 | 98.48 | 62.00 | 0.97 | 0.24 | 98.46 | 64.47 | 98.48 | 61.82 |
| ECMLAU01 | 98.24 | 64.52 | 0.97 | 0.29 | 98.21 | 65.61 | 98.24 | 64.48 |
| ECMLAU02 | 98.72 | 71.16 | 0.97 | 0.42 | 98.71 | 71.03 | 98.72 | 71.16 |
| ECMLBU00 | 97.00 | 64.75 | 0.94 | 0.30 | 96.91 | 62.60 | 97.00 | 64.63 |
| ECMLBU01 | 96.25 | 70.25 | 0.93 | 0.41 | 96.10 | 67.75 | 96.24 | 70.07 |
| ECMLBU02 | 96.75 | 84.75 | 0.94 | 0.70 | 96.73 | 84.32 | 96.75 | 84.74 |
| ECMLBU03 | 97.75 | 84.75 | 0.96 | 0.70 | 97.76 | 84.71 | 97.75 | 84.75 |
| ECMLBU04 | 96.50 | 77.50 | 0.93 | 0.55 | 96.55 | 78.16 | 96.50 | 77.48 |
| ECMLBU05 | 93.25 | 81.50 | 0.87 | 0.64 | 93.20 | 81.42 | 93.25 | 81.74 |
| ECMLBU06 | 93.50 | 71.00 | 0.87 | 0.43 | 93.56 | 68.13 | 93.50 | 70.76 |
| ECMLBU07 | 97.50 | 74.75 | 0.95 | 0.52 | 97.45 | 70.38 | 97.50 | 74.19 |
| ECMLBU08 | 98.25 | 74.50 | 0.97 | 0.49 | 98.22 | 72.87 | 98.25 | 74.41 |
| ECMLBU09 | 95.50 | 80.50 | 0.91 | 0.61 | 95.29 | 80.40 | 95.49 | 80.50 |
| ECMLBU10 | 93.75 | 80.00 | 0.88 | 0.60 | 93.70 | 78.84 | 93.75 | 79.94 |
| ECMLBU11 | 91.50 | 74.50 | 0.84 | 0.49 | 90.96 | 72.58 | 91.47 | 74.37 |
| ECMLBU12 | 97.00 | 69.00 | 0.94 | 0.38 | 96.97 | 68.04 | 97.00 | 68.97 |
| ECMLBU13 | 97.50 | 78.00 | 0.95 | 0.56 | 97.47 | 78.54 | 97.50 | 77.99 |
| ECMLBU14 | 94.50 | 76.75 | 0.89 | 0.54 | 94.33 | 76.57 | 94.50 | 76.75 |
| PU1 | 97.97 | 96.75 | 0.96 | 0.93 | 97.63 | 96.20 | 97.93 | 96.68 |
| PU2 | 96.18 | 94.27 | 0.88 | 0.81 | 89.41 | 83.02 | 93.54 | 89.79 |
| PU3 | 97.57 | 95.50 | 0.95 | 0.91 | 97.22 | 95.02 | 97.53 | 95.52 |
| PUA | 96.13 | 95.20 | 0.92 | 0.90 | 96.19 | 95.29 | 96.12 | 95.20 |

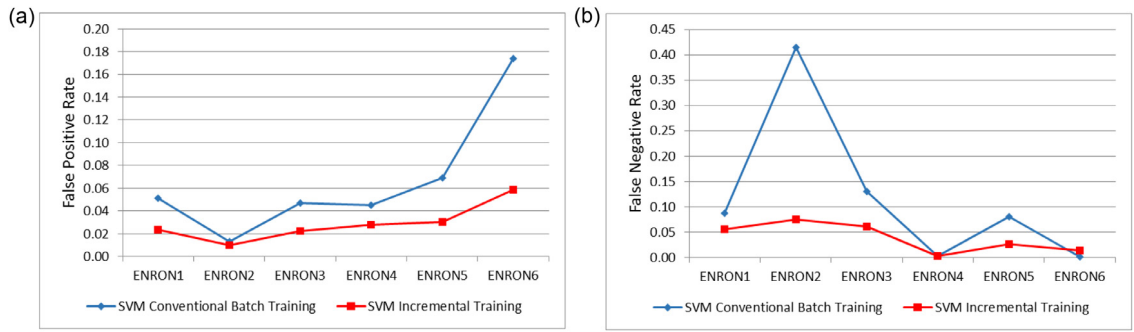

**Fig. 6.** Results of Enron dataset (a) false positive rate (b) false negative rate.

### 4.4. Discussion

In this paper, we have presented a complete e-mail spam filtering system that is capable of updating its prediction ability according to the nature of incoming data. We analyzed the classification results to compare and effectively validate the performance of the proposed feature selection function. The e-mail classification with batch learning model was conducted to evaluate TFDCR function and to examine filter performance in the presence of concept drift. Enron dataset includes e-mails sorted as per their arrival time. Each folder of Enron is partitioned into three subsets from which the first subset is used for the training purpose, and rest of the two are further partitioned into ten subsets to create ten different testing instances. We observed the increase in the misclassification rate as the number of testing samples is increasing in the batch learning mode. The e-mail spam filter performance in the case ECML dataset is somewhat degraded in the batch mode because the training data is collected from publicly available combined sources. The test data is taken from the inbox of several individual users resulting in a completely different distribution. There is also a vast difference in the training set size in ECML-A dataset and ECML-B dataset. However, PU dataset includes personalized folders with different size and proportion of spam and legitimate e-mails.

The significant amount of research work has been carried out in the literature to train SVM using fast and simplified algorithms that demand fewer memory resources. Our SVM classification model uses Sequential Minimal Optimization (SMO) algorithm (Platt, 1999) for the filter implementation. SMO is a special case of decomposition method wherein each sub-problem, two coeffi-cients are optimized per iteration which is solved analytically to improve the objective function. It operates by solving the smallest possible optimization problem at every step. The SMO algorithm selects two Lagrange multipliers to optimize jointly, finds the optimal values for these multipliers, and updates the SVM to reflect the new optimal values. A set of support vectors comprises those training vectors that explicitly determine the optimal decision hyperplane. An attractive approach towards incremental learning process involves optimizing a new set of examples with support vectors that lead to the reduction in problem dimensionality. The proposed framework is tested for e-mail classification problem, and significant computational benefits are achieved by avoiding the re-training from scratch. In addition, the incremental learning framework enables the classifier to be dynamically adaptive in the presence of drifting concepts.

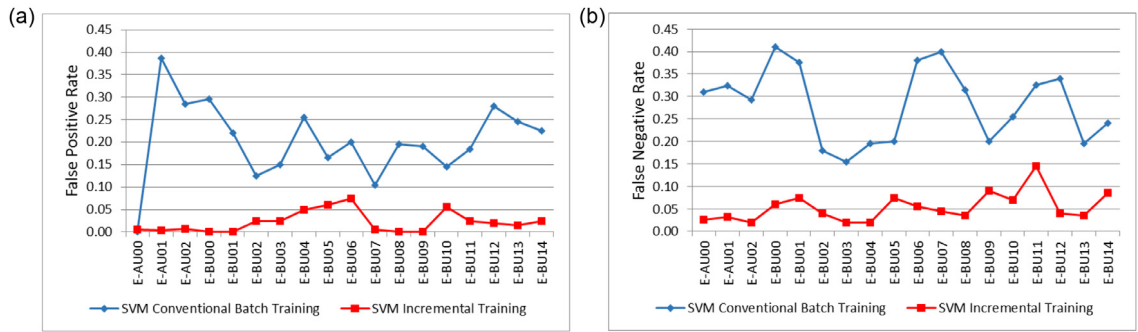

Fig. 7. Results of ECML dataset (a) false positive rate (b) false negative rate.

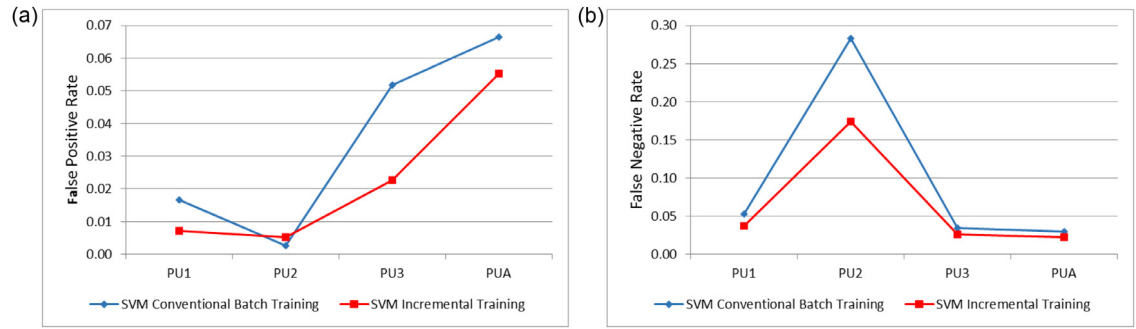

Fig. 8. Results of PU dataset (a) false positive rate (b) false negative rate.

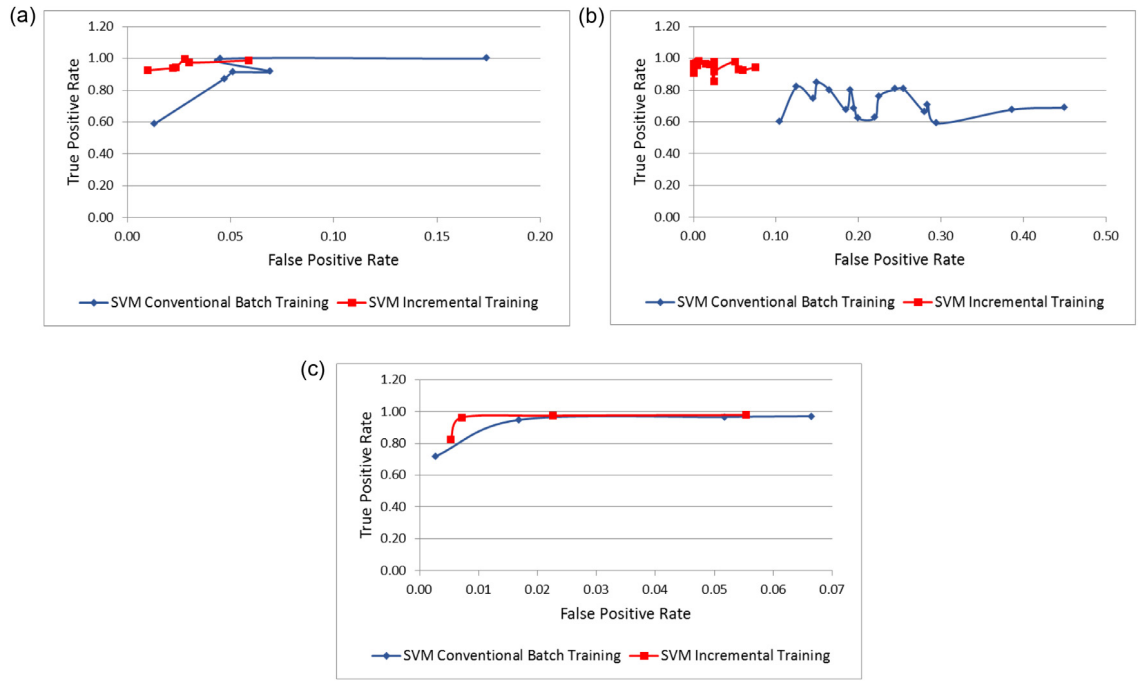

Fig. 9. ROC curve (a) Enron dataset (b) ECML dataset. (c) ROC curve PU dataset.

The performance of classifier directly validates the discriminative power of feature space created by a specific feature selection function. We estimated that a feature carries higher discriminating ability in the case when it has a considerable difference of occurrence in the different classes. To address the issue of uneven class distribution, we utilized the value of feature's category ratio of the class in which its presence is dominating and the reverse category ratio where it occurs less frequently. The simple formulation of the proposed TFDCR function requires less computational time for its operations. For the Enron and PU datasets, the performance of TFDCR is comparable with IG and chi-square feature selection functions. However, there is no single function, which shows the superiority. IG involves the estimation of conditional probabilities of a class given an attribute and entropy. It ignores the correlation between attributes. Chi-square calculates the statistics between every feature and the target class and observes the existence of a relationship between the feature and the class. It determines the divergence from the expected distribution assuming the independence between feature occurrence and the class value. Gini index measures the impurity of a feature, i.e., the ability of a feature to distinguish between the possible classes. Gain ratio modifies the information gain to reduce the effect of the most common terms and the marginal probabilities of the term by normalizing their weights. CFS takes into account the individual predictive ability of each feature along with the degree of redundancy between them. Correlation coefficient estimates the correlation between a subset of features and class and inter-correlations between the features. In the case of ECML dataset, where the distribution shift is present in the e-mail data, TFDCR outperforms the rest of the functions for all 18 folders except the two.

Furthermore, to enhance the performance of the classifier, we integrated the incremental learning framework with dynamic feature update function. We use the accuracy and false positive rate as the indicators for activating incremental learning. The noticeable advantage of our approach is, it allows the incremental up gradation of the existing feature set and capacitates the classifier to re-learn the decision model according to new samples using modified feature set. However, the essential constraint of our incremental learning framework is, the user feedback is needed to provide the true labels of previously misclassified e-mails and the e-mails of the testing set which violated the constraint in order to re-train the classifier. With each incremental pass, we observed 3 to –5 percent of update took place in the existing feature set by the proposed *selectionRankWeight* function.

## 5. Conclusion and future work

Personalized e-mail spam filtering has been one of the challenging classification tasks in the presence of concept drift. In this paper, we have presented the design and development of e-mail spam filter that incorporates an effective term frequency difference and category ratio based novel feature selection function TFDCR, an incremental learning model using SVM to update the classifier dynamically and a feature update function. The experimental results demonstrate the distinctive theoretical and practical implications in the concerned domain. This study showed the successful applicability of the proposed incremental learning framework using SVM with the integration of dynamic feature update function. Results obtained from experiment 2 confirmed the role of support vectors and a small set of additional vectors to update the learned decision hypothesis. The potential benefit of it is re-training the decision model using the old solution as a starting point and saving the resources which would otherwise require for re-training the classifier from scratch. Another significant contribution of this paper is novel proposed TFDCR feature selection function for binary classification problem. TFDCR feature selection function combines feature's term frequency difference & class category ratio and generates a subset of the most discriminative features. The computation of TFDCR is straightforward. The discriminative weight measuring parameters of TFDCR formula can be easily derived using simple arithmetic operations. The function has been tested on three different benchmark datasets consisting of total 28 folders of personalized e-mails. Experimental results prove the efficacy of our function to be applied for selecting features in text classification problems.

Some of the improvements and extensions help to explore various research directions and can be executed in the future. They are discussed below:

- Our study is focused on the development of personalized e-mail spam filter that works as an extended model for classification of e-mails at the client side. Further enhancement can be done to make it semi-personalized for a group of users belonging to the same organization or an institution. A separate filter then learns in a collaborative environment from the collection of e-mails for each such group to make the classification decision.
- The filter is incrementally updated when the performance measures violate the threshold value. Currently, the user-defined threshold is used and separated for each user inbox. The threshold value can be learned automatically for each different category. Also, a provision can be made to choose a higher threshold for the categories that are more prone to distribution shift. However, it is likely to increase the performance but will also incur an overhead due to frequent incremental training.
- The incremental learning model is enhanced using our proposed *selectionRankWeight* function to upgrade the existing feature set. An automatic intelligent detection system can be incorporated which continually monitors the evolution of new features tend to have higher discriminating ability from incoming samples and stores them for the future update. Such an approach may save time which is currently spent for explicit recognition which is taking place in the existing algorithm.
- The proposed TFDCR feature selection function is derived for binary classification problems. It can be extended to work for multi-class classification problems as well. Our function considers the difference of term occurrence in both the categories which can be replaced as the difference between total occurrence and a specific category occurrence. The other term is a category ratio which would be independent of the number of categories.

To conclude, in this paper we considered the adversarial characteristics of e-mail classification problem such as concept drift, uneven class distribution, different uncertain misclassification error cost and the requirement of personalized dynamically adaptable e-mail spam filter. Our incremental learning framework using SVM overcomes the limitation of a static decision model to handle the concept drift and substantially improves the performance of the classifier. Our proposed system is well generalized, robust and can be used as a state of the art classifier for binary text classification in different application domains especially where a system requires being adaptable to the dynamically changing conditions.

### Acknowledgment

We are thankful to the Nirma University, India for providing resources and other facilities to carry out this research work. This research did not receive any specific grant from funding agencies in the public, commercial, or not-for-profit sectors.

# References

Agnihotri, D., Verma, K., & Tripathi, P. (2017). Variable global feature selection scheme for automatic classification of text documents. *Expert Systems with Applications, 81*, 268–281.

Androutsopoulos, I., Koutsias, J., Chandrinos, K. V., & Spyropoulos, C. D. (2000). An experimental comparison of naive bayesian and keyword-based anti-spam filtering with personal e-mail messages. In *Proceedings of the 23rd annual international ACM SIGIR conference on research and development in information retrieval* (pp. 160–167). ACM. doi:10.1145/345508.345569.

Breiman, L., Friedman, J., Stone, C., & Olshen, R. (1984). *Classification and regression trees*.

Chang, M.-w., Yih, W.-t., & Mccann, R. (2008). Personalized spam filtering for gray mail. In *Proceedings of the fifth conference on email and anti-spam (CEAS)*.

Cheng, V., & Li, C. H. (2006). Personalized spam filtering with semi-supervised classifier ensemble. In *Proceedings of the 2006 IEEE/WIC/ACM international conference on web intelligence* (pp. 195–201).

Cormack, G. V. (2008). Email spam filtering: a systematic review. *Foundations and Trends in Information Retrieval, 1*, 335–455.

Cortes, C., & Vapnik, V. (1995). Support-vector networks. *Machine Learning, 20*, 273–297.

Delany, S. J., Cunningham, P., Tsymbal, A., & Coyle, L. (2005). A case-based technique for tracking concept drift in spam filtering. *Knowledge-Based Systems, 18*, 187–195.

Discovery Challenge. Retrieved 4 2016, from http://www.ecmlpkdd2006.org/challenge.html.

Drucker, H., Wu, D., & Vapnik, V. N. (1999). Support vector machines for spam categorization. *IEEE Transactions on Neural networks, 10*, 1048–1054.

E-mail Statistics Report, 2013-2017. (n.d.). Retrieved 5 from https://www.radicati.com/wp/wp-content/uploads/2013/04/E-mail-Statistics-Report-2013-2017-Executive-Summary.pdf.

Enron Spam Data sets. Retrieved 3 2016, from http://csmining.org/index.php/enron-spam-datasets.html.

Fawcett, T. (2003). In vivo spam filtering: A challenge problem for KDD. *ACM SIGKDD Explorations Newsletter, 5*, 140–148.

Fdez-Riverola, F., Iglesias, E. L., Diaz, F., Mendez, J. R., & Corchado, J. M. (2007). Applying lazy learning algorithms to tackle concept drift in spam filtering. *Expert Systems with Applications, 33*, 36–48.

Gama, J., Zliobaite, I., Bifet, A., Pechenizkiy, M., & Bouchachia, A. (2014). A survey on concept drift adaptation. *ACM Computing Surveys (CSUR), 46*, 44.

Georgala, K., Kosmopoulos, A., & Paliouras, G. (2014). Spam filtering: An active learning approach using incremental clustering. In *Proceedings of the 4th international conference on web intelligence, mining and semantics (WIMS14)* (p. 23).

Gray, A., & Haahr, M. (2004). *Personalised, Collaborative Spam Filtering*.

Hall, M. A. (2000). Correlation-based feature selection of discrete and numeric class machine learning. In *Proceedings of the seventeenth international conference on machine learning*.

Hsiao, W.-F., & Chang, T.-M. (2008). An incremental cluster-based approach to spam filtering. *Expert Systems with Applications, 34*, 1599–1608.

Joachims, T. (1998). Text categorization with support vector machines: Learning with many relevant features. In *European conference on machine learning* (pp. 137–142).

Junejo, K. N., & Karim, A. (2013). Robust personalizable spam filtering via local and global discrimination modeling. *Knowledge and Information Systems, 34*, 299–334.

Jung, J., & Sit, E. (2004). An empirical study of spam traffic and the use of DNS black lists. In *Proceedings of the 4th ACM SIGCOMM conference on internet measurement* (pp. 370–375).

Karegowda, A. G., Manjunath, A. S., & Jayaram, M. A. (2010). Comparative study of attribute selection using gain ratio and correlation based feature selection. *International Journal of Information Technology and Knowledge Management, 2*, 271–277.

Katakis, I., Tsoumakas, G., & Vlahavas, I. (2006). Dynamic feature space and incremental feature selection for the classification of textual data streams. *Knowledge Discovery from Data Streams*, 107–116.

Katakis, I., Tsoumakas, G., & Vlahavas, I. (2010). Tracking recurring contexts using ensemble classifiers: An application to email filtering. *Knowledge and Information Systems, 22*, 371–391.

Laskov, P., Gehl, C., Kruger, S., & Muller, K.-R. (2006). Incremental support vector learning: Analysis, implementation and applications. *Journal of Machine Learning Research, 7*, 1909–1936.

Lee, C., & Lee, G. G. (2006). Information gain and divergence-based feature selection for machine learning-based text categorization. *Information Processing & Management, 42*, 155–165.

Matthews, B. W. (1975). Comparison of the predicted and observed secondary structure of T4 phage lysozyme. *Biochimica et Biophysica Acta (BBA)-Protein Structure, 405*, 442–451.

Michelakis, E., Androutsopoulos, I., Paliouras, G., Sakkis, G., & Stamatopoulos, P. (2004). Filtron: A learning-based anti-spam filter. *proceedings of the 1st conference on email and anti-spam. Mountain*.

Pinheiro, R. H., Cavalcanti, G. D., & Ren, T. I. (2015). Data-driven global-ranking local feature selection methods for text categorization. *Expert Systems with Applications, 42*, 1941–1949.

Platt, J. C. (1999). 12 fast training of support vector machines using sequential minimal optimization. *Advances in Kernel Methods*, 185–208.

Rao, J. M., & Reiley, D. H. (2012). 9). The Economics of Spam. *Journal of Economic Perspectives, 26*, 87–110.

Santos, I., Laorden, C., Sanz, B., & Bringas, P. G. (2012). Enhanced topic-based vector space model for semantics-aware spam filtering. *Expert Systems with Applications, 39*, 437–444.

Shams, R., & Mercer, R. E. (2013). Personalized spam filtering with natural language attributes. In *Machine learning and applications (ICMLA), 2013 12th international conference on: 2* (pp. 127–132).

Shih, D.-H., Chiang, H.-S., & Lin, B. (2008). Collaborative spam filtering with heterogeneous agents. *Expert Systems with Applications, 35*, 1555–1566.

Shilton, A., Palaniswami, M., Ralph, D., & Tsoi, A. C. (2005). Incremental training of support vector machines. *IEEE Transactions on Neural Networks, 16*, 114–131.

Syed, N. A., Liu, H., & Sung, K. K. (1999). Handling concept drifts in incremental learning with support vector machines. In *Proceedings of the fifth ACM SIGKDD international conference on knowledge discovery and data mining* (pp. 317–321).

Teng, W.-L., & Teng, W.-C. (2008). A personalized spam filtering approach utilizing two separately trained filters. In *Proceedings of the 2008 IEEE/WIC/ACM international conference on web intelligence and intelligent agent technology* (pp. 125–131). 02.

Uysal, A. K. (2016). An improved global feature selection scheme for text classification. *Expert Systems with Applications, 43*, 82–92.

Uysal, A. K., & Gunal, S. (2014). The impact of preprocessing on text classification. *Information Processing & Management, 50*, 104–112.

Wang, Y., Liu, Y., Feng, L., & Zhu, X. (2015). Novel feature selection method based on harmony search for email classification. *Knowledge-Based Systems, 73*, 311–323.

Widmer, G., & Kubat, M. (1996). Learning in the presence of concept drift and hidden contexts. *Machine Learning, 23*, 69–101.

Yang, Y., & Pedersen, J. O. (1997). A comparative study on feature selection in text categorization. *Icml, 97*, 412–420.

Yevseyeva, I., Basto-Fernandes, V., Ruano-OrdaS, D., & MeNdez, J. R. (2013). Optimising anti-spam filters with evolutionary algorithms. *Expert Systems with Applications, 40*, 4010–4021.

Youn, S., & McLeod, D. (2009). Spam decisions on gray e-mail using personalized ontologies. In *Proceedings of the 2009 ACM symposium on applied computing* (pp. 1262–1266).